\documentclass[twoside]{article}

\usepackage[accepted]{aistats2022}
\usepackage[round]{natbib}
\usepackage[utf8]{inputenc} 
\usepackage[T1]{fontenc}    
\usepackage{hyperref}       
\usepackage{url}            
\usepackage{booktabs}       
\usepackage{amsfonts}       
\usepackage{nicefrac}       
\usepackage{xcolor}         
\usepackage{lipsum}

\usepackage{graphicx}
\usepackage[ruled, vlined]{algorithm2e} 
\usepackage{subfigure}
\usepackage{float}
\usepackage{amsmath}

\usepackage{mathtools}

\DeclarePairedDelimiter\floor{\lfloor}{\rfloor}

\newcommand{\OurModel}{DGHL}
\newcommand{\OurModelsp}{DGHL }

\begin{document}

\twocolumn[

\aistatstitle{Deep Generative model with Hierarchical Latent Factors for Time Series Anomaly Detection}

\aistatsauthor{Cristian Challu\footnotemark[1] \And Peihong Jiang \And  Ying Nian Wu \And Laurent Callot }

\aistatsaddress{ CMU \And  Amazon Research \And Amazon Research and UCLA \And Amazon Research} ]

\footnotetext[1]{Work performed while at AWS AI Labs. }

\begin{abstract}

Multivariate time series anomaly detection has become an active area of research in recent years, with Deep Learning models outperforming previous approaches on benchmark datasets. Among reconstruction-based models, most previous work has focused on Variational Autoencoders and Generative Adversarial Networks. This work presents DGHL, a new family of generative models for time series anomaly detection, trained by maximizing the observed likelihood by posterior sampling and alternating back-propagation. A top-down Convolution Network maps a novel hierarchical latent space to time series windows, exploiting temporal dynamics to encode information efficiently. Despite relying on posterior sampling, it is computationally more efficient than current approaches, with up to 10x shorter training times than RNN based models. Our method outperformed current state-of-the-art models on four popular benchmark datasets. Finally, DGHL is robust to variable features between entities and accurate even with large proportions of missing values, settings with increasing relevance with the advent of IoT. We demonstrate the superior robustness of DGHL with novel occlusion experiments in this literature. Our code is available at \url{https://github.com/cchallu/dghl}.

\end{abstract}

\setcounter{footnote}{2}
\section{INTRODUCTION}\label{section:intro}

\begin{figure}[ht!]
\centering
\includegraphics[width=\linewidth]{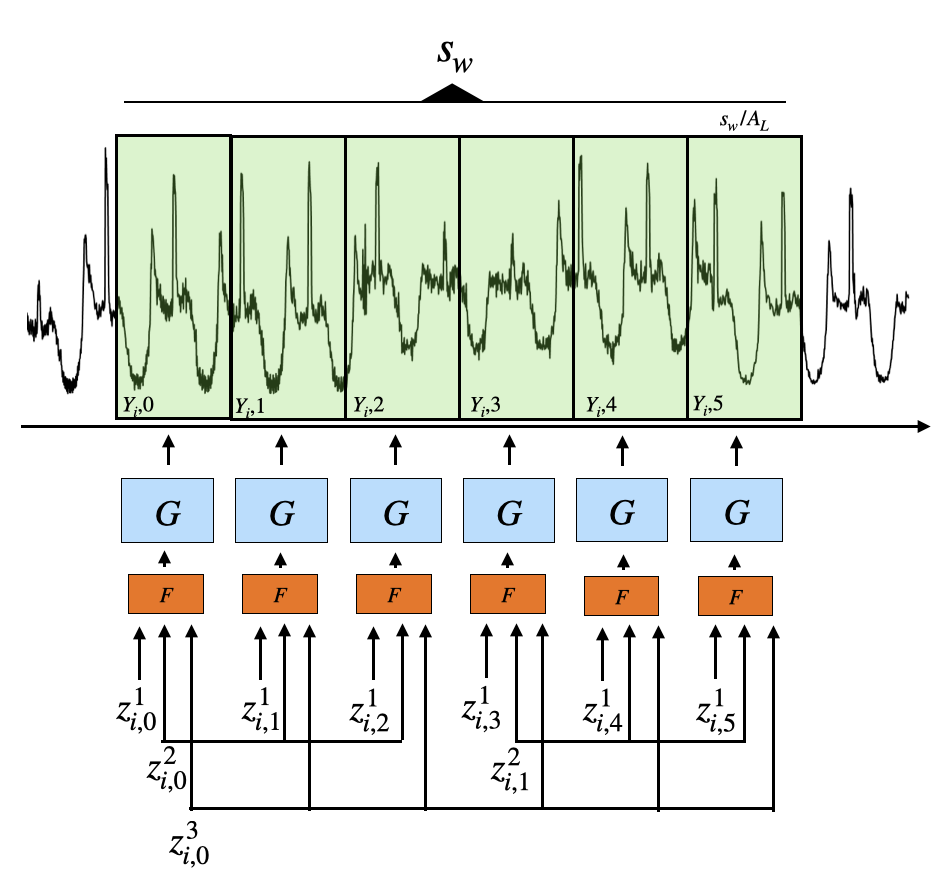} 
\caption{Example of hierarchical latent factor space for $\boldsymbol{a} = [1,3,6]$. On each level $l$, the latent vectors of $a_l$ consecutive sub-windows are tied. For instance, the latent vector on the highest layer, $L$, is shared by all sub-windows of $\boldsymbol{Y}$.}
\label{fig:model}
\end{figure}

Recent advancements in Deep Learning such as Recurrent Neural Networks (RNN), Temporal Convolution Networks (TCN) and Graph Networks (GN) have been successfully incorporated by recent models to outperform previous approaches such as out-of-limits, clustering-based, distance-based, and dimensionality reduction on a wide range of tasks. \citep{chalapathy2019deep} \citep{ruff2021unifying} present comprehensive reviews of current state-of-the-art methods for time-series anomaly detection.

In this work, we propose \OurModel, a novel Deep Generative model based on a top-down Convolution Network (ConvNet), which maps multivariate time-series windows to a novel hierarchical latent space. The model is trained by maximizing the observed likelihood directly with the Alternating Back-Propagation algorithm \citep{han2017alternating} and short-run MCMC \citep{nijkamp2021learning}, so it does not rely on auxiliary networks such as encoders or discriminators as VAEs and GANs do. \OurModel, therefore, comprehends a separate family of generative models, previously unexplored for time-series anomaly detection. We perform experiments on several popular datasets and show the proposed model outperforms the recent state-of-the-art while reducing training times against previous reconstruction-based and generative models.

With the advent of IoT, we believe that settings with corrupted or missing data have increasing relevance. For example, faulty sensors can cause missing values, privacy issues on consumer electronics devices, or heterogeneous hardware can lead to variable features. We present the first extensive analysis on the robustness of current state-of-the-art models on datasets with missing inputs and variable features with novel occlusion experiments. \OurModelsp achieved superior performance on this setting, maintaining state-of-the-art performance with up to 90\% of missing data, without modification to the architecture or training procedure. We perform additional qualitative experiments of our model to assess desirable properties of lower-dimensional representations such as continuity and extrapolation capabilities. Finally, we show how \OurModelsp can be used as a forecasting model, demonstrating its versatility on various time-series tasks.   

The main \textbf{contributions} of our paper are:

\begin{itemize}
	\item \textbf{Short-run MCMC}. First time series anomaly detection generative model based on short-run MCMC for estimating posterior of latent variables and inferring latent vectors. In particular, first application of Alternating Back-Propagation algorithm for learning generative model for time-series data.
	\item \textbf{Hierarchical latent factors}. We present a novel hierarchical latent space representation to generate windows of arbitrary length. We demonstrate with ablation studies how \OurModelsp achieves state-of-the-art performance by leveraging this representation on four benchmark datasets.
	\item \textbf{Robustness to missing data}. We present the first experiments on robustness to missing inputs of state-of-the-art anomaly detection models, and demonstrate \OurModelsp achieves superior performance in this setting.
	\item \textbf{Open-source implementation}. We publish an open implementation of our model and full experiments for reproducibility of the results.
\end{itemize}

The rest of the paper is structured as follows. Section \ref{section:related} reviews current state-of-the-art models, Section \ref{section:model} introduces \OurModelsp and describes the Alternating Back-Propagation algorithm, Section \ref{section:experiments} contains the empirical findings. In section \ref{section:discussion} we present a discussion of the main findings. Finally in Section \ref{section:conclusion} we wrap up and conclude.

\section{RELATED WORK}\label{section:related}


\subsection{Reconstruction-based models}

Reconstruction-based models learn representations for the time-series by reconstructing the input based on latent variables. The reconstruction error or the likelihood are commonly used as anomaly scores. Among these models, variational auto-encoders (VAE) are the most popular. The LSTM-VAE, proposed in \citep{park2018multimodal}, uses LSTM both as encoders and decoders and models the reconstruction error with support vector regression (SVR) to have a dynamic threshold based on the latent space vector. OmniAnomaly \citep{su2019robust} improves on the LSTM-VAE by adding normalizing planar flows to increase the expressivity and including a dynamic model for the latent space.

Generative Adversarial Networks (GANs) were also adapted for anomaly detection as alternatives to VAE, with models such as AnoGAN \citep{schlegl2017unsupervised}, MAD-Li \citep{li2018anomaly}, and MAD-GAN \citep{li2019mad}. For instance, in MAD-GAN, a GAN is used to generate short windows of time-series with LSTM Generator and Discriminator networks. The anomaly score considers both the reconstruction error of the reconstructed window by the Generator network and the score of the Discriminator network.

MTAD-GAT \citep{zhao2020multivariate} proposed to combine both forecasting and reconstruction approaches. It includes a Multi Layer Perceptron (MLP) for forecasting and a VAE for reconstructing the time-series, with the anomaly score including both forecasting and reconstruction loss terms. It also includes two Graph Attention Networks (GAT) to model temporal dynamics and correlations explicitly. \citep{deng2021graph} later proposed GDN, which models the relationships between time-series with a Graph Neural Network and uses a GAT for forecasting. \\

Most recent models propose to detect anomalies directly on the latent representation and embeddings. THOC \citep{shen2020timeseries} proposed to use one-class classifiers based on multiple hyperspheres on the representations on all intermediate layers of a dilated RNN. NCAD \citep{carmona2021neural} uses a TCN to map context windows and suspect windows into a neural representation and detect anomalies in the suspect window on the latent space with a contextual hypersphere loss.


\subsection{Generative models with Alternating Back-Propagation}

Virtually all current models, including our proposed approach, rely on mapping the original time-series input into embeddings or a lower-dimensional latent space. \OurModel, however, is trained with the Alternating Back-Propagation (ABP) algorithm, presented in \citep{han2017alternating}, and short-run MCMC, presented in \citep{nijkamp2021learning}. ABP maximizes the observed likelihood directly; it does therefore not rely on variational inference approximations or auxiliar networks such as discriminators. Instead, our approach uses MCMC sampling methods to sample from the true posterior to approximate the likelihood gradient.

Several generative models which rely on MCMC sampling, and in particular Langevin Dynamics, have shown state-of-the-art performance on computer vision \citep{pang2020learning} and NLP \citep{pang2021generative} tasks. To our knowledge, this algorithm has not been used for time-series forecasting and time-series anomaly detection. We present the ABP algorithm in more detail in subsection \ref{section:abp}.

\section{\OurModel}\label{section:model}

\subsection{Hierarchical Latent Factors}

\begin{figure*}[ht!]
\centering
\includegraphics[width=0.8\linewidth]{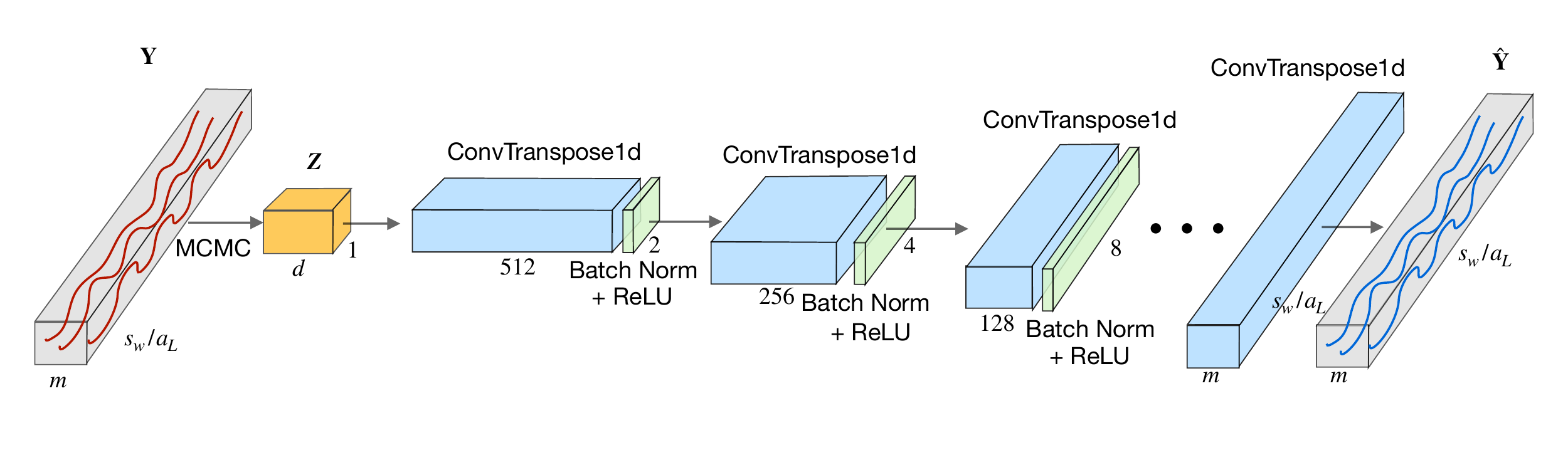} 
\caption{\textit{Generator} architecture, which maps a latent vector $\boldsymbol{Z}$ to a time-series window. Each layer of the ConvNet increases the temporal dimension and reduces the filters by a factor of 2. A batch normalization layer and ReLU activations are included between each convolutional layer.}
\label{fig:convnet}
\end{figure*}

Let $\boldsymbol{Y} \in \mathbb{R}^{m \times s_w}$ be a window of size $s_w$ of a multivariate time-series with $m$ features. The window $\boldsymbol{Y}$ is further divided in sub-windows of equal length $\boldsymbol{Y_j} \in \mathbb{R}^{m \times \frac{s_w}{a_L}}, j=0,...,a_{L}$. The structure of the hierarchy is specified by $\boldsymbol{a}=[a_1,...,a_{L}]$, where $L$ is the number of levels, and $a_l$ determines the number of consecutive sub-windows with shared latent vector on level $l$, with $a_{l} \mid a_L$ . Our model for each sub-window $\boldsymbol{Y_j}$ of $\boldsymbol{Y}$ is given by,

\begin{equation}\label{eq:model}
\begin{split}
\boldsymbol{s_j} &= F_{\boldsymbol{\alpha}}(\boldsymbol{z}^1_{\floor{\frac{j}{a_1}}}, ..., \boldsymbol{z}^L_{\floor{\frac{j}{a_L}}}) \\
\boldsymbol{Y_j} &= G_{\boldsymbol{\beta}}(\boldsymbol{s_j}) + \boldsymbol{e_j}
\end{split}
\end{equation}

where $F_\alpha$ is the \textit{State model}, $G_\beta$ is the \textit{Generator model}, $\boldsymbol{\theta}=[{\boldsymbol{\alpha}}, {\boldsymbol{\beta}}]$ are the parameters, $\boldsymbol{s_j} \in \mathbb{R}^d$ is the state vector, $\boldsymbol{e_j} \sim N(0,\boldsymbol I_D)$, and 
\begin{equation}
\boldsymbol{Z}=\{\boldsymbol{z}^l_{\floor{\frac{j}{A_l}} } \in \mathbb{R}^{d_l} \}_{l,j}
\end{equation}

is the hierarchical latent factor space for window $\boldsymbol{Y}$. For the \textit{State model} we used a concatenation layer. For the \textit{Generator model} we used a top-down Convolution Network (ConvNet), shown in Figure \ref{fig:convnet}, which maps an input state vector to a multivariate time-series window. The \textit{State model} for each sub-window has $L$ latent vectors inputs. On each level $l$, the latent vectors of $a_l$ consecutive sub-windows are tied. For instance, the latent vector on the highest layer, $L$, is shared by all sub-windows of $\boldsymbol{Y}$. Figure 1 shows an example of a hierarchical latent space with $\boldsymbol{a} = [1,3,6]$.

The key principle of the hierarchical latent space is to leverage dynamics on the time-series, such as seasonalities, to encode the information on the latent space more efficiently, with lower-dimensional vectors. The hierarchical latent space allows generating realistic time-series of arbitrary length while preserving their long-term dynamics. The hierarchical structure can be incorporated as hyper-parameters to be tuned or pre-defined based on domain knowledge. For instance, hierarchies can correspond to the multiple known seasonalities on the time-series.

The hierarchical latent space $\boldsymbol{Z}$ is jointly inferred using Langevin Dynamics. The relative size of the lowest level state vector and the upper levels controls the flexibility of the model. Larger lower hierarchy level vectors make the model more flexible, making it robust to normal changes or randomness in long-term dependencies of the time-series and therefore reducing false positives by reducing the reconstruction error. Larger tied vectors will make the model more strict, better for detecting contextual anomalies. The model presented in \citep{han2017alternating} can be seen as a single level hierarchical latent space model, with $\boldsymbol{a}=[1]$, in the current framework.

Previous work such as OmniAnomaly incorporates transition models to learn dynamics in the latent space. We believe our proposed hierarchical latent factors structure has several advantages over transition models. First, the computational cost and training time is lower for the proposed model since it does not rely on sequential computation and therefore on back-propagation through time for training parameters. Second, transition models implicitly assume the dynamics are constant over time, a non-realistic assumption in many settings. Our solution allows the model to share information across windows to model long-term dynamics without relying on a parametric model which assumes constant dynamics.

\subsection{Training with Alternating Back-Propagation}\label{section:abp}

The parameters $\boldsymbol{\theta}$ of \OurModelsp are learned with the Alternating Back-Propagation algorithm. First, the training multivariate time-series $\boldsymbol{Y} \in \mathbb{R}^{m \times T}$ with $m$ features and $T$ timestamps, is divided in consecutive windows of size $s_w$ and step size $s$ in a rolling-window fashion. Let $\{\mathbf{Y}^{(i)}, i=1,...,n\}$ be the training set of time-series windows. Alternating Back-Propagation algorithm learns parameters $\boldsymbol{\theta}$ by maximizing the observed log-likelihood directly, given by,
\begin{equation}
	L(\boldsymbol{\theta}) = \sum_{i=1}^{n} \log p_{\boldsymbol{\theta}}(\mathbf{Y}^{(i)}) = \sum_{i=1}^{n} \log \int p_{\boldsymbol{\theta}}(\mathbf{Y}^{(i)},\mathbf{Z}^{(i)}) d \mathbf{Z}^{(i)}
\end{equation}

where $\mathbf{Z}^{(i)}$ is the latent vector for window $i$ specified in equation 2. The observed likelihood $L(\boldsymbol{\theta})$ is analytically intractable. However, the gradients $L'(\boldsymbol{\theta})$ for a particular observation can be simplified to,

\begin{equation}
\begin{split}
\frac{\partial}{\partial \boldsymbol{\theta}} \log p_{\boldsymbol{\theta}} (\mathbf{Y}^{(i)}) &= \frac{1}{p_{\boldsymbol{\theta}}(\mathbf{Y}^{(i)})}\frac{\partial}{\partial \boldsymbol{\theta}} \int p_{\boldsymbol{\theta}}(\mathbf{Y}^{(i)},\mathbf{Z}^{(i)}) d \mathbf{z} \\
& = \mathbb{E}_{p_{\boldsymbol{\theta}}(\mathbf{Z}^{(i)}|\mathbf{Y}^{(i)})}\left[ \frac{\partial}{\partial \boldsymbol{\theta}} \log p_{\boldsymbol{\theta}} (\mathbf{Y}^{(i)},\mathbf{Z}^{(i)})\right]
\end{split}
\end{equation}

where $p_{\boldsymbol{\theta}}(\mathbf{Z}^{(i)}|\mathbf{Y}^{(i)}) = p_{\boldsymbol{\theta}}(\mathbf{Y}^{(i)},\mathbf{Z}^{(i)})/p_{\boldsymbol{\theta}}(\mathbf{Y}^{(i)})$ is the posterior. The expectation in the previous equation can be approximated with the Monte Carlo average by taking samples using MCMC. In particular, Alternating Back-Propagation takes a single sample of the posterior using Langevin Dynamics \citep{neal2011mcmc}, a Hamiltonian Monte Carlo algorithm, which iterates,

\begin{equation}\label{eq:langevin}
\begin{split}
\mathbf{Z}^{(i)}_{t+1} &= \mathbf{Z}^{(i)}_t + \frac{s}{\sigma_z} \frac{\partial}{\partial \mathbf{Z}^{(i)}} \log p_{\boldsymbol{\theta}}(\mathbf{Z}^{(i)}_t|\mathbf{Y}^{(i)}) + \sqrt{2s} \epsilon_t \\
&= \mathbf{Z}^{(i)}_t + \sqrt{2s}\epsilon_t + \\
 & \frac{s}{\sigma_z}\left[ (\mathbf{Y}^{(i)}-f(\mathbf{Z}^{(i)}_t,\boldsymbol{\theta}))\frac{\partial}{\partial \mathbf{Z}^{(i)}}f(\mathbf{Z}^{(i)}_t,\boldsymbol{\theta})-\mathbf{Y}^{(i)}_t\right]
\end{split}
\end{equation}

where $\boldsymbol{\epsilon}_t \sim N(0,\boldsymbol{I}_D)$, $t$ is the time step of the dynamics, $s$ is the step size, and $\sigma_z$ controls the relative size of the injected noise. This iteration is an explain-away process where latent factors are chosen such that the current residual on the reconstruction, $\mathbf{Y}^{(i)}-f(\mathbf{Z}^{(i)}_t,\boldsymbol{\theta})$, is minimized. With large values of $\sigma_z$, the posterior will be close to the prior, while small $\sigma_z$ allows for a richer posterior. The iterative process is truncated to a predefined number of iterations, and the rejection step is not considered. As explained in \citep{neal2011mcmc}, for an observation $\mathbf{Y}^{(i)}$, the resulting vector is a sample from an approximated posterior, $p_{\boldsymbol{\theta}}(\mathbf{Z}^{(i)}|\mathbf{Y}^{(i)})$. The Monte Carlo approximation of the gradient then becomes,

\begin{equation} \label{eq:gradient}
\begin{split}
L'(\boldsymbol{\theta}) &\approx \frac{\partial}{\partial \boldsymbol{\theta}} \log p_{\boldsymbol{\theta}}(\mathbf{Z}^{(i)},\mathbf{Y}^{(i)}) \\
& = \frac{1}{\sigma^2} (\mathbf{Y}^{(i)} - f(\mathbf{Z}^{(i)},\boldsymbol{\theta})) \frac{\partial}{\partial \boldsymbol{\theta}} f(\mathbf{Z}^{(i)},\boldsymbol{\theta})
\end{split}
\end{equation}

Algorithm 1 presents the alternating back-propagation algorithm with mini-batches. Analogous to the EM algorithm \citep{dempster1977maximum}, it iterates two distinct steps: (1) \textit{inferential back-propagation} and (2) \textit{learning back-propagation}. During (1), the latent vectors $\{\mathbf{Z}^{(i)}\}$ are inferred for a sample $\{\mathbf{Y}^{(i)}, i=1,..., b_s \}$. In step (2), $\{\mathbf{Z}^{(i)}\}$ are used as input of the Generator model $f$ and parameters $\boldsymbol{\theta}$ are updated with SGD. We use Adam optimizer for learning parameters with default parameters \citep{kingma2014adam}.

\begin{algorithm}
\SetAlgoLined
\SetKwInOut{Input}{input}\SetKwInOut{Output}{output}
\Input{training examples $\{\mathbf{Y}^{(i)}, i=1,...,n\}$, Langevin steps $l$, learning iterations T}
\Output{learned parameters $\boldsymbol{\theta}$, inferred latent vectors $\{\mathbf{Z}^{(i)}, i = 1,...,n\}$}
 Let $t \leftarrow 0$, initialize $\boldsymbol{\theta}$ \\
 Initialize $\mathbf{Z}^{(i)}$, for $i=1,...,n$ \\
 \While{$t<T$}{
 Take a random mini-batch $\{\mathbf{Y}^{(j)}, j=1,...b\}$. \\
 \textbf{Inferential back-propagation:} For each $j$, run $l$ steps of Langevin dynamics to sample $\mathbf{Z}^{(j)} \sim p(\mathbf{Z}|\mathbf{Y}^{(j)}, \boldsymbol{\theta})$ following equation \ref{eq:langevin}. \\
 \textbf{Learning back-propagation:} Compute gradients following equation \ref{eq:gradient} and update $\boldsymbol{\theta}$. \\
 $t \leftarrow t + 1$
 }
 \caption{Mini-batch Alternating back-propagation}
\end{algorithm}

One disadvantage of MCMC methods is the computational cost. Langevin Dynamics, however, relies on the gradients of the Generator function, which can be efficiently computed with modern automatic differentiation packages such as Tensorflow and PyTorch. Moreover, back-propagation on ConvNet is easily parallelizable in GPU. Several recent works, have shown models trained with alternating back-propagation \citep{han2017alternating}, \citep{xie2019learning} and short-run MCMC \citep{nijkamp2021learning} achieved state-of-the-art performance in computer vision and NLP tasks while remaining computationally efficient and comparable in training time to methods relying solely on SGD. We discuss training and inference times on section \ref{section:experiments}.

As described in \citep{welling2011bayesian}, Bayesian posterior sampling provides inbuilt protection against overfitting. The MCMC sampling allows \OurModelsp to model complex multivariate time-series, while reducing the risk of overfitting. This is particularly helpful on problems with small training sets.

\subsection{Online Anomaly Detection}

By learning how to generate time-series windows based on the training data $\boldsymbol{Y}$, \OurModelsp implicitly learns \textit{normal} (non-anomalous) temporal dynamics and correlations between the multiple time-series. In this subsection, we explain the proposed approach to reconstruct windows on unseen test data $\boldsymbol{Y}^{test}$ to detect anomalies.

In Online Anomaly Detection we consider the test set $\boldsymbol{Y}^{test} \in \mathbb{R}^{m \times T_{test}}$ to be a stream of $m$ time-series. The goal is to detect anomalies (the evaluation is equivalent to a supervised setting with two classes) as soon as possible. As with the training set, $\boldsymbol{Y}^{test}$ is first divided in consecutive windows with the same parameters $s_w$ and $s$. We propose to reconstruct and compute anomalies scores one window at a time.

Let $\boldsymbol{Y_{t^*}}$ be the current window of interest. The latent space $\boldsymbol{Z_{t^*}}$ is  jointly inferred to reconstruct the target window, namely $\boldsymbol{\hat{Y}_{t^*}}$. The anomaly score for a particular timestamp $t$ in the window is computed as the Mean Square Error (MSE) considering all $m$ time-series, given by

\begin{equation}
s_t = \frac{1}{m}\sum_{i=1}^m(y_{i,t}-\hat{y}_{i,t})^2
\end{equation}

The size of the window $s_w$ and step size $s$ control how early anomalies can be detected. With a smaller $s$, anomaly scores for newer values in the stream are computed sooner. When $s < s_w$, consecutive windows have overlapping timestamps. In this case, scores are updated by considering the average reconstruction. In datasets with multiple entities (for instance, machines in SMD), we scale the scores by the accumulated standard deviation of scores before window $t^*$.
 
One main difference with the inference step during training is the removal of the Gaussian noise, $\epsilon_t$, of the Langevin Dynamics update. The inferred factors then correspond to the maximum a posteriori mode, which in turn minimizes the reconstruction error conditional on the learned models $F$ and $G$. This novel strategy makes \OurModelsp unique among reconstruction-based models: it avoids overfitting during training by sampling from the posterior with Langevin Dynamics and minimizes the reconstruction error to reduce false positives by MAP estimation.

Many previous models rely on complex and unusual specific scores, but \OurModelsp uses the simple MSE. The anomaly scores of our approach are interpretable, since they can be disaggregated by the $m$ features. Users can rank the contribution to the anomaly score of each feature to gather insights of the anomaly.

\subsection{Online Anomaly Detection with missing data}
 
The first step of the ABP algorithm is to infer latent vectors with Langevin Dynamics. This is an explain-away process where latent factors are chosen such that the current residual on the reconstruction, $\mathbf{Y}-f(\mathbf{Z}_t,\boldsymbol{\theta})$, is minimized. The model can intrinsically deal with missing data by inferring $\mathbf{Z}$ computing the residuals only on the observed signal $\mathbf{Y}_{obs}$. The inferred vectors then correspond to samples from the posterior distribution conditional on the available signal, $p_{\boldsymbol{\theta}}(\mathbf{Z}|\mathbf{Y}_{obs})$. Since no explicit learnable parameters map inputs to the latent space, the model is more robust to missing values and outliers (masked as missing data). Generative models trained with ABP algorithm outperformed VAEs and GANs on experiments with missing information on computer vision and NLP tasks \citep{han2017alternating}.

\begin{figure}[ht!]
\centering
\includegraphics[width=\linewidth]{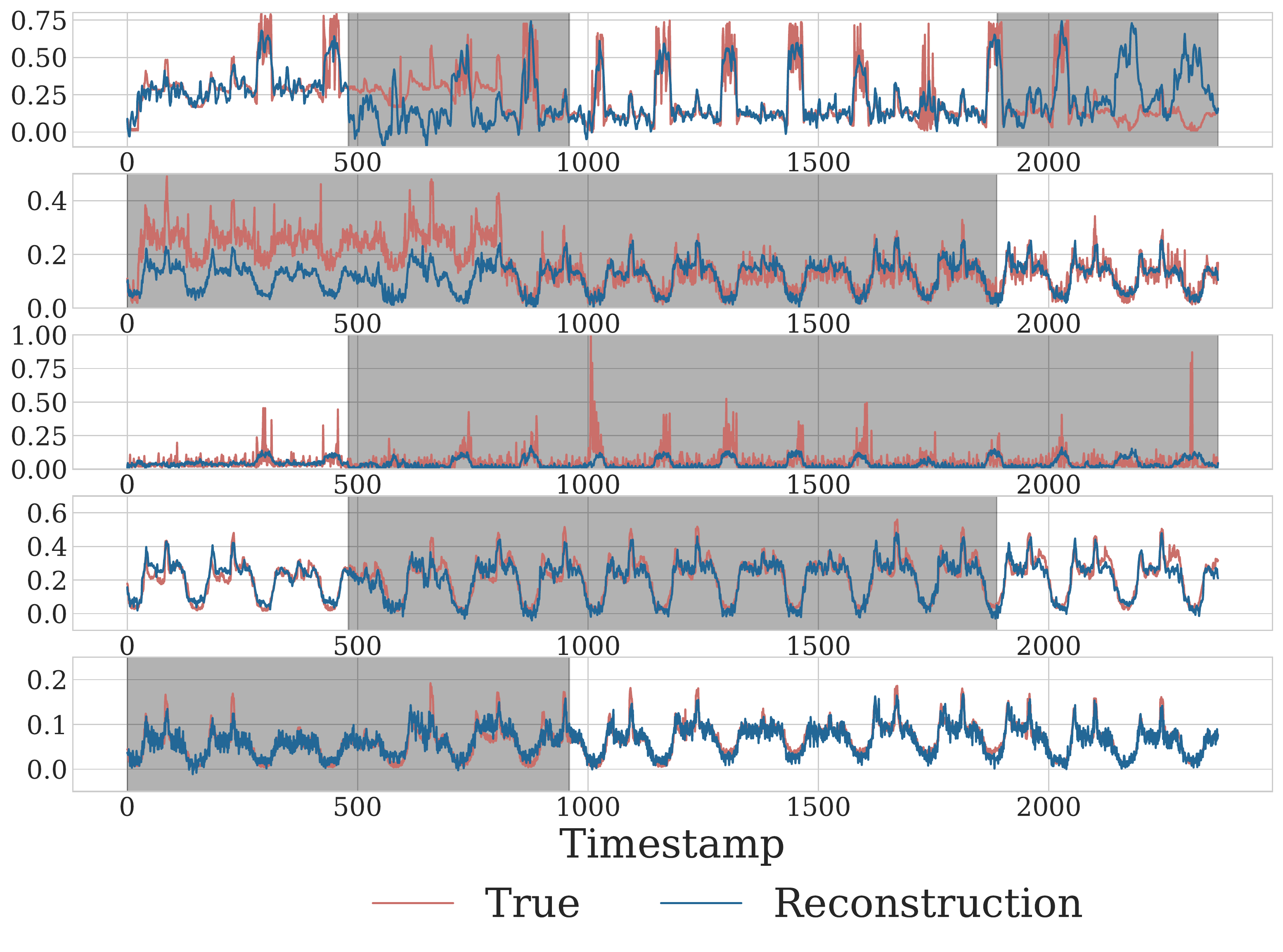} 
\caption{Occlusion experiment on machine-1-1 of the SMD. Red lines correspond to the actual values and blue lines present the reconstructed time-series with \OurModel. Gray areas correspond to the occluded information during training.}
\label{fig:occluded}
\end{figure}

Figure \ref{fig:occluded} shows an example of occluded data, for a subset of features of one machine of the SMD dataset. Occluded segments are marked in gray. First, \OurModelsp is able to precisely reconstruct the observed data (white region), even when most features are missing. This is most relevant for the anomaly detection task, since only the observed features are used to compute the anomaly score. Second, the model is able to recover missing data with great precision, which can be helpful in complete pipelines with downstream applications.

\section{EXPERIMENTS}\label{section:experiments}

\subsection{Datasets}
\textbf{Server Machine Dataset (SMD)} $-$ Introduced in \citep{su2019robust}, SMD is a multivariate time-series dataset with 38 features for 28 server machines, monitored during 5 weeks. The time-series include common activity metrics in servers such as CPU load, network and memory usage, among others. Both training and testing sets contain around 50k timestamps each, with 5 \% of anomalous cases. We trained separate models for each machine as suggested by the authors but with the same hyperparameters. \\

\textbf{Soil Moisture Active Passive satellite (SMAP) and Mars Science Laboratory rover (MSL)$-$} Published by NASA in \citep{hundman2018detecting}, they contain real telemetric data of the SMAP satellite and MSL rover. SMAP includes 55 multivariate time-series datasets, each containing one anonymized channel and 24 variables encoding information sent to the satellite. MSL includes 27 datasets, each with one telemetry channel and 54 additional variables. Again, we train separate models for each telemetry channel, considering additional variables as exogenous, ie. only the anomaly score of the telemetry channel was used for detecting anomalies. \\

\textbf{Secure Water Treatment (SWaT)} $-$ Is a public dataset with information of a water treatment testbed meant for cyber-security and anomaly detection research. It contains network traffic and data from 51 sensors for 11 days, 7 days of normal operation (train set) and 4 days with cyber attacks (test set).

\subsection{Evaluation}

We evaluate the performance of \OurModelsp and benchmark models on the four datasets with the $F_1$-score, considering the anomaly detection problem as a binary classification task where the positive class corresponds to anomalies. Anomalies often occur continuously over a period of time creating anomalous segments. \citep{xu2018unsupervised} proposed an adjustment approach, where the predicted output is re-labeled as an anomaly for the whole continuous anomalous segment if the model correctly identifies the anomaly in at least one timestamp. We use this adjustment technique for SMAP, MSL and SMD datasets to make results comparable with existing literature. Moreover, we followed the common practice of comparing the performance using the best $F_1$-score, by choosing the best threshold on the test set. For SMAP, MSL and SMD we use a single threshold through the entire dataset (not different thresholds for each machine or channel).

\begin{figure}[ht!]
\centering
\includegraphics[width=\linewidth]{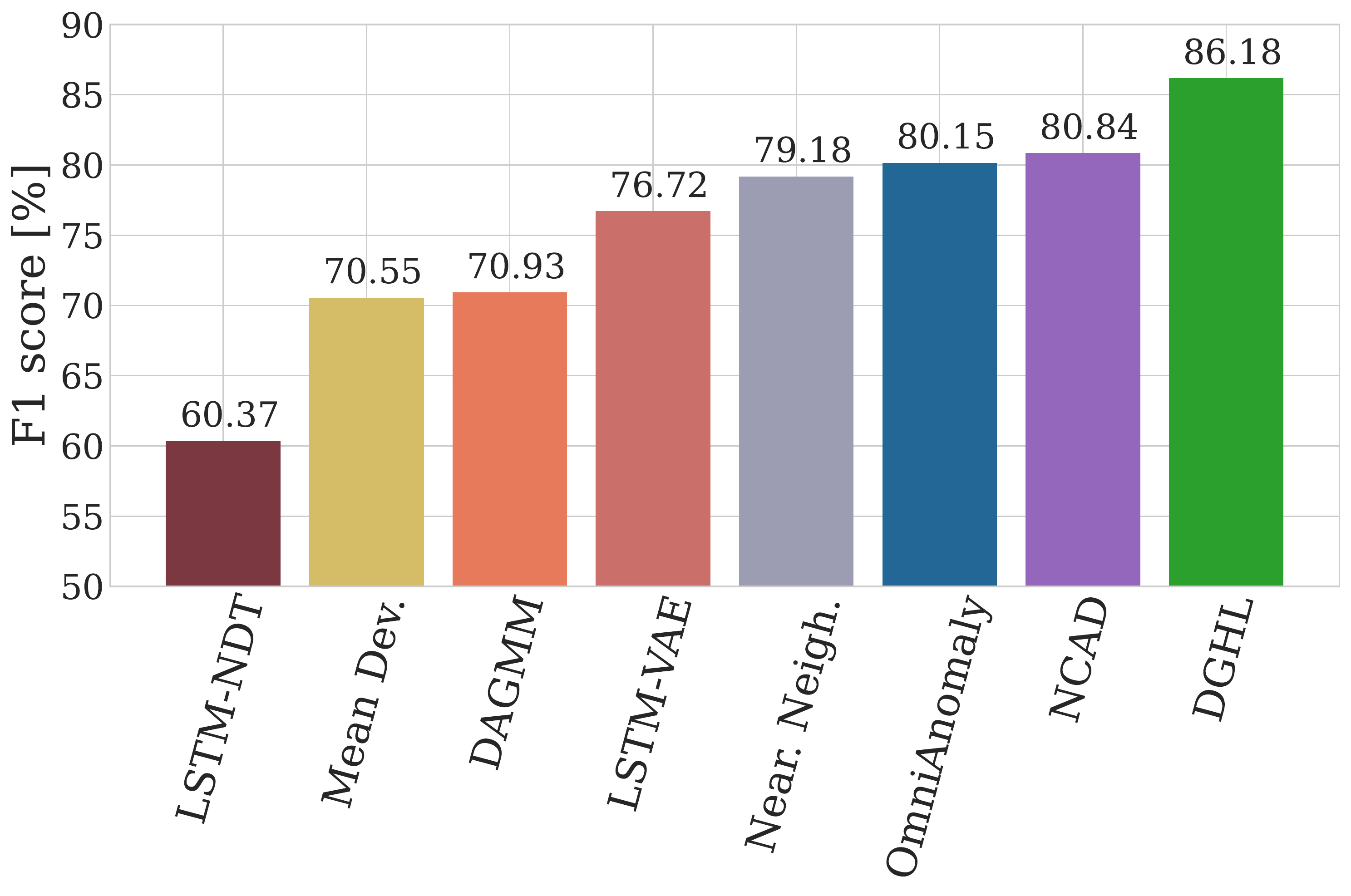} 
\caption{$F_1$ scores on SMD dataset using a single threshold across all machines.}
\label{fig:f1_smd}
\end{figure} 

To make our results comparable with previous work, we follow the train, validation and test split described in \citep{shen2020timeseries} for SMAP, MSL and SWaT. For SMD we use the train and test splits described in \citep{su2019robust}. All architecture hyper-parameters of the \textit{Generator} model, optimization hyper-parameters, and all hyper-parameters of the Langevin Dynamics were kept constant across the four datasets.

We compare \OurModelsp to current state-of-the art models, such as THOC \citep{shen2020timeseries}, NCAD \citep{carmona2021neural}, and MTAD-GAT \citep{zhao2020multivariate}; and previous widely used models such as AnoGAN \citep{schlegl2017unsupervised}, DeepSVDD \citep{ruff2018deep}, DAGMM \citep{zong2018deep}, OmniAnomaly, \citep{su2019robust}, MAD-GAN \citep{li2019mad} and LSTM-VAE \citep{park2018multimodal}. We also include simple one-line and non deep-learning approaches such as \textit{Mean deviation} and \textit{Nearest Neighbors}. \textit{Mean deviation} uses the average deviation to the mean of each feature as anomaly score. The later uses the average distance to the k nearest windows of the training set as anomaly score. Finally, we include two additional versions of \OurModelsp removing the key contributions of our work. First, we remove the hierarchical factors ($\mathbf{a}=[1])$, and second, we replace the Langevin Dynamics algorithm for inferring latent factors with a convolutional encoder.

Table \ref{table:main_result} shows the $F_1$ scores for \OurModel, and the benchmark models for SMAP, MSL, and SWaT datasets. Our methods consistently achieves the Top-2 $F_1$ scores, with overall performance superior to state-of-the-art such as MTAD-GAT \citep{zhao2020multivariate}, THOC and NCAD. Moreover, our approach achieved the highest performance between all reconstruction based and generative models on all datasets. Figure 2 shows the $F_1$ scores for SMD dataset. \OurModelsp achieved a score of $86.18 \pm 0.66$, outperforming all benchmark models. Our model without hierarchical factors had a score of $80.84 \pm 0.40$, and with encoder had a score of $76.59 \pm 0.78$. \footnotetext{MTAD-GAT authors do not provide public implementation of the model nor evaluation on SWaT. NCAD results on SWaT is not comparable since they use segment adjustment.}

\begin{table*}[h]
  \caption{$F_1$ scores on benchmark datasets (the larger the better). The benchmark models performance was taken from \citep{shen2020timeseries}, \citep{carmona2021neural} and \citep{zhao2020multivariate}. First place is marked in bold and second place in bold and italic. \OurModelsp corresponds to the full model described in previous section, without Hierarchical factors corresponds to the simpler model with fully independent latent vectors for each window.}
  \label{table:main_result}
  \centering
  \begin{tabular}{lccc}
    \toprule
    \textbf{Model} & \textbf{SMAP}  & \textbf{MSL} & \textbf{SWaT} \\
    \midrule
    Mean deviation (one-line) & 57.61 & 68.91 & 85.71 \\
    Nearest Neighbors & 75.10 & 90.01 & 86.72 \\
    \midrule
    AnoGAN & 74.59 & 86.39 & 86.64 \\
    DeepSVDD &  71.71 & 88.12 & 82.82  \\   
    DAGMM & 82.04 & 86.08 & 85.37 \\
    LSTM-VAE & 75.73 & 73.79 & 86.39 \\
    MAD-GAN & 81.31 & 87.47 & 86.89 \\
    MSCRED & 85.97 & 77.45 & 86.84 \\
    OmniAnomaly & 85.35 & 90.14 & 86.67 \\
    MTAD-GAT & 90.13 & 90.84 & - \\
    THOC & \textit{\textbf{95.18}} & 93.67 & \textbf{88.09} \\
    NCAD & 94.45 & \textbf{95.60} & - \\
    \midrule
    \OurModel & \textbf{96.38 $\pm$ 0.72} & \textit{\textbf{94.08}} $\pm$ \textit{\textbf{0.35}} & \textit{\textbf{87.47 $\pm$ 0.22}} \\
    \OurModel, without Hierarchial factors & 94.87 $\pm$ 0.71  & 91.26 $\pm$ 0.71 & 87.08 $\pm$ 0.12 \\
    \OurModel, with encoder, no Hier. factors & 78.41 $\pm$ 0.92  & 87.10 $\pm$ 0.54 & 86.39 $\pm$ 0.32 \\
    \bottomrule
  \end{tabular}
\end{table*}

\OurModelsp significantly outperformed simple baselines in all datasets. The one-line solution ranked worst consistently. Nearest Neighbors, however, achieved a better performance than several complex models in all datasets with a fraction of the computational cost, demonstrating how simple models need to be considered to understand the benefits of recent models. \OurModelsp outperforms other pure reconstruction-based models because inferring latent vectors for computing anomaly scores provides several advantages. First, it provides additional flexibility and generalization capabilities to prevent false positives, which is instrumental in noisy or non-constant temporal dynamics datasets. Second, it helps to reduce the lasting impact of anomalies on the reconstruction error over time, reducing false positives once anomalies end.

\OurModelsp took an average of 2 minutes to train for each entity (e.g. one machine of SMD or one chanel of SMAP) consistently across datasets. For instance, the training time was around 60 minutes for SMD and MSL, and 100 minutes for SMAP. This is comparable to other state-of-the-art models self-reported training times such as NCAD and faster than RNN based models. For instance, OmniAnomaly took an average of 20 minutes to train each model for each machine on the SMD dataset. The inference time varies depending on the length of the test set. The average time to infer 3000 timestamps (average downsampled SMD test set), with $s_w=32$, was lower than 5 seconds.

\subsection{Online Anomaly Detection with missing data}

All current benchmark datasets in the time-series anomaly detection literature assume \textit{perfect} data. However, this is not usually the case in real scenarios, with issues like missing values, corrupted data, and variable features. This section presents the first experiments to assess the robustness of current state-of-the-art models to common data issues such as missing values. In particular, we adapt the popular occlusion experiments from computer vision literature for training models with incomplete data. 

\begin{figure}[ht!]
\centering
\includegraphics[width=\linewidth]{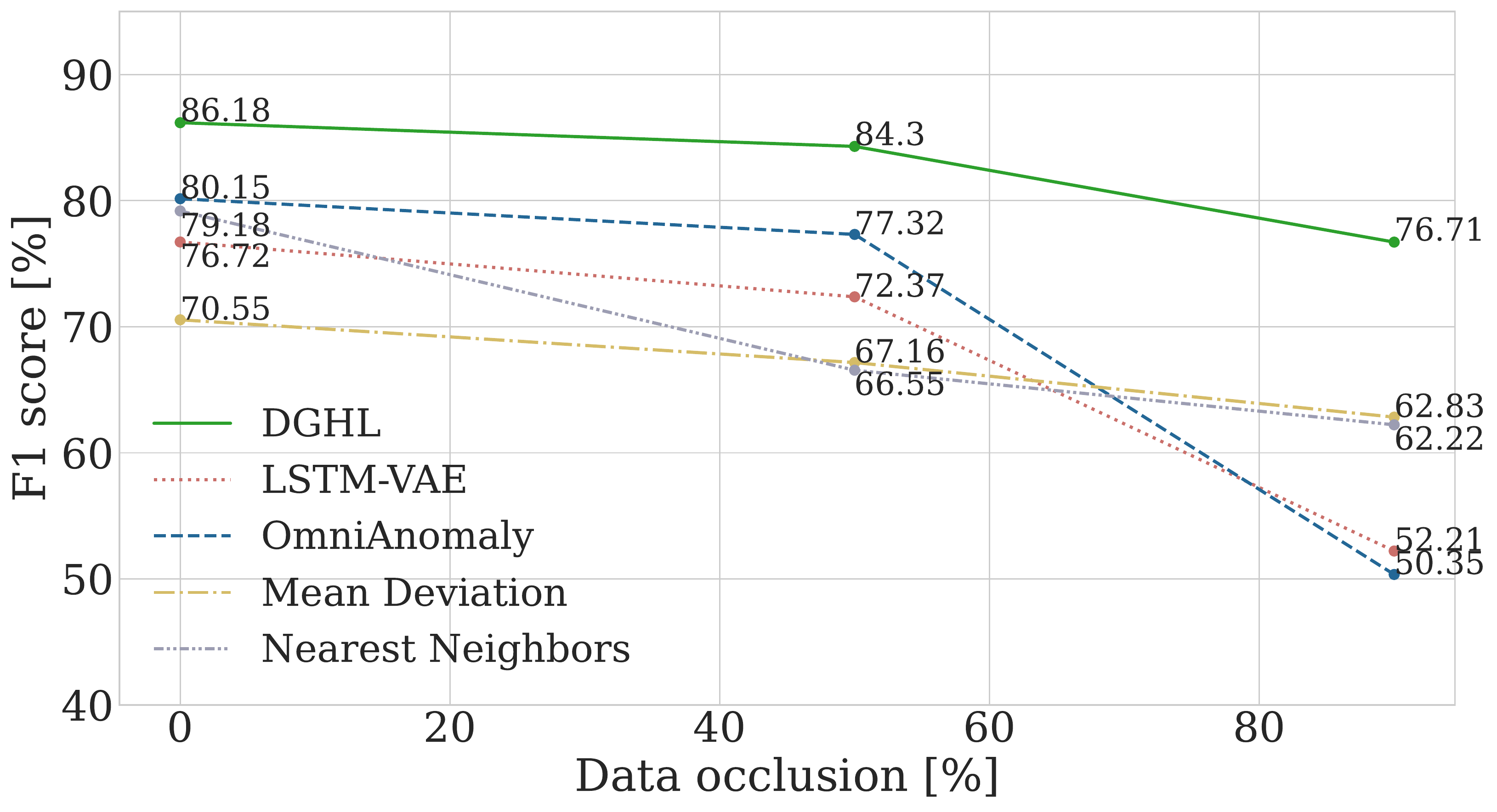} 
\caption{$F_1$ scores for \OurModelsp and LSTM-VAE benchmark for occlusion experiments on SMD for three levels of occlusion probability $p$, 0, 0.5, 0.9 and $r=5$.}
\label{fig:f1_occluded}
\end{figure} 

We define the occlusion experiments with two parameters. First, the original time-series $\boldsymbol{Y} \in \mathbb{R}^{m \times T}$, is divided in $r$ segments of equal length, $\boldsymbol{Y_i} \in \mathbb{R}^{m \times \frac{T}{r}}$. Second, each feature $m$ in each segment is occluded for model training or inference with probability $p$.

We assess the robustness of models to incomplete training data with occluding experiments on the SMD dataset, for different levels of $r$ and $p$, and using $F_1$ scores to evaluate performance.  Figure \ref{fig:f1_occluded} shows the $F_1$ score for \OurModelsp, LSTM-VAE, and OmniAnomaly. \OurModelsp achieves the highest scores consistently, with an increasing relative performance on higher data occlusion probability. Moreover, \OurModelsp maintained high $F_1$ scores even with up to 90\% of missing information, without any changes to the hyperparameters, architecture or training procedure.

\subsection{Time-series generation}

\OurModelsp is trained to generate time-series windows from a latent space representation. We examine how \OurModelsp learns the representation by interpolating and extrapolating between two latent vectors, $\boldsymbol{Z}_l$ and $\boldsymbol{Z}_u$, inferred from two windows of a real time-series from the SMD. The trained \textit{Generator network} is then used to generate new windows across the interpolation subspace.

\begin{figure*}[ht!]
\centering
\includegraphics[width=0.8\linewidth]{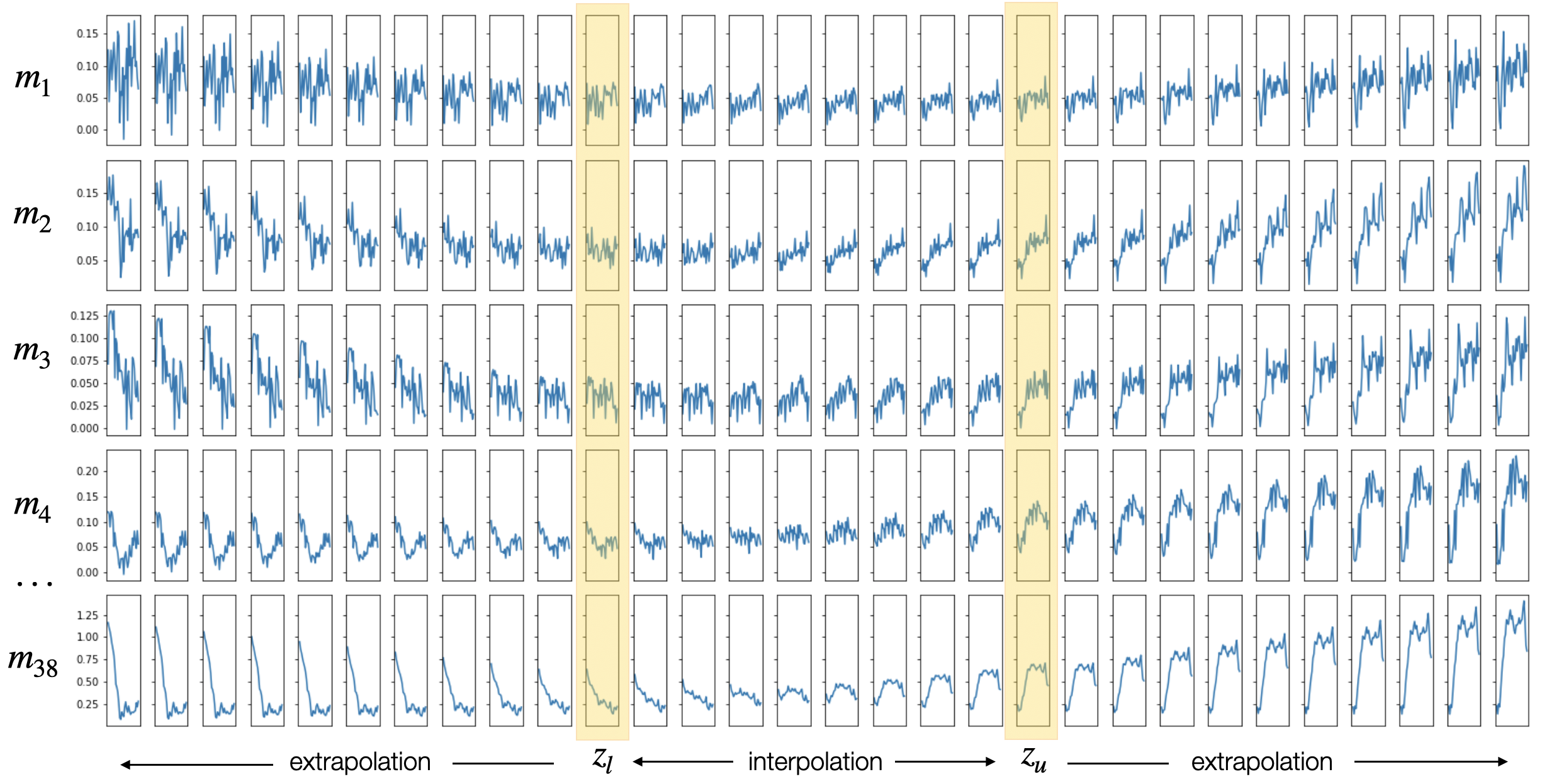} 
\caption{Generated time-series windows from interpolation and extrapolation of latent vectors using \OurModelsp trained on machine-1-1 of the SMD.}
\label{fig:representation}
\end{figure*}

Figure \ref{fig:representation} presents the generated windows for interpolated and extrapolated vectors for a subset of the original features. The generated time-series smoothly transition between clear patterns on both shape and scale. Moreover, \OurModelsp is able to generate meaningful time-series on the extrapolation region. This experiment shows how our approach maps similar time-series windows into close points of the latent space, which is a desirable property of latent representations.

\section{DISCUSSION}\label{section:discussion}

\OurModelsp outperforms SoTA baselines and simple approaches by the current experimental standards of the literature. Although \OurModelsp relies on MCMC for posterior sampling, it remains computationally efficient thanks to a lower number of training iterations needed. The ablations studies presented in Table \ref{table:main_result} demonstrates the complementary gains of our two main contributions, namely, a novel hierarchical latent representation and training the \textit{Generator} with the alternating back-propagation algorithm. 

\citep{wu2020current} strongly argue that some of the benchmark datasets used in our experiments have mislabeled observations, and are therefore worthless to compare to. While we agree this adds noise to the evaluation metrics, we believe the consistent improvement of our model (and current SoTA deep learning models) demonstrates their superior performance on this task over simpler approaches. We also observed the SMD dataset does not have a considerable amount of mislabel observations that could significantly alter the results \footnote{Even though we do not know the ground truth, most labeled anomalies seem to relate to some form of rare pattern, and we did not observe clear anomalies (e.g. large spikes) labeled as normal observations.}. We decided to use these benchmark datasets for comparison purposes with existing methods.

Most of the anomalies in the benchmark datasets used in this paper can be easily identified by humans (e.g. large spikes), as noted in \citep{wu2020current}. Accurately detecting such anomalies is relevant for many applications, in particular large-scale automatized systems, for which the current benchmarks provide representative estimates of the relative performance of models. Identifying contextual anomalies, which can be hard to detect even for humans, is also highly relevant but current benchmarks do not provide insights on the performance of models on such tasks. Creating relevant benchmark datasets with contextual anomalies is a pressing necessity.

As shown in equation \ref{eq:model}, our approach assumes a constant variance. The normalization by the accumulated standard deviation of scores help to account for the variance of each future, allowing to have single thresholds across different entities. It would be interesting to extend \OurModelsp to also model the variance of each feature. We show in the supplementary material how \OurModelsp can be used for forecasting. An interesting research question is to evaluate how this approach performs on the forecasting task. The hierarchical latent factors allows the model to reconstruct long time-series, by learning long-term temporal relations, making our approach useful on multivariate long-horizon forecasting \citep{zhao2020multivariate, challu2022n}. Moreover, as demonstrated with both NASA datasets, \OurModelsp can incorporate exogenous variables, which are crucial in several applications such as electricity price forecasting \citep{olivares2021neural}.
 
We do not believe the paper's contribution can be directly misused to have a negative societal impact. We recommend performing additional thorough experiments on healthcare before using the proposed approach in applications in this domain. \OurModelsp required significantly less computational resources and training time than most current state-of-the-art models, but it still uses high-performance hardware and more resources than simpler models. These additional costs should be considered on applications compared to the improved performance benefits.

\section{CONCLUSION}\label{section:conclusion}

In this paper, we introduced \OurModel, a state-of-the-art Deep Generative model for multivariate time-series anomaly detection. The proposed model maps time-series windows to a novel hierarchical latent space representation, which leverages the time-series dynamics to encode information more efficiently. A ConvNet is used as the \textit{Generator}. \OurModelsp does not rely on auxiliary networks, such as encoders or discriminators; instead, it is trained by maximizing the likelihood directly with the Alternating Back-Propagation algorithm. Our model has several advantages over existing methods: i. shorter training times, ii. demonstrated superior performance on several benchmark datasets, and iii. better robustness to missing values and variable features.

\section*{Aknowledgments}\label{section:aknowledgments}
We thank Kin Olivares and François-Xavier Aubet for insightful conversations.

\clearpage
\bibliography{citations.bib}

\clearpage
\appendix
\thispagestyle{empty}
\twocolumn[\makesupplementtitle]


\section{Hyperparameters}

The hyperparameters of our method are divided in the following groups:
\begin{itemize}
	\item \textit{Generator} architecture: number of convolutional filters, window size, hierarchical structure and step size.
	\item MCMC: latent space dimension and Langevin Dynamics update parameters.
	\item Optimization: learning rate, decay, batch size and number of iterations.
\end{itemize}

The hyperparameters where selected based on reconstruction error from a validation set of the SMD (constructed from the training set to maintain comparability). For the Langevin Dynamics hyperparameters we used the values proposed in previous work, which are known to work across many different datasets and applications.

We use the same hyperparameters for the four benchmark datasetes, SMAP, MSL, SWAT and SMD. This demonstrates the versatility of our method and how it can potentially be used in real datasets with minimal hyperparameter tuning, further reducing its computational cost. The following table shows the final hyperparameters used in the experiments.

\begin{table}[!ht] 
\caption{Hyperparameters}
\label{table:hyperparameters}
\centering
    \begin{tabular}{ll}
    \toprule
    \textsc{Hyperparameter}                               & \textsc{Value}     \\ \midrule
    Window size ($s_w/a_L$).                              & $64$               \\
    Hierarchical structure ($\mathbf{a}$).                & $[1, 4]$           \\
    Step size ($s$).									  & $256$              \\
    ConvNet filters multiplier.                        	  & $32$               \\ 
    Max filters per layer.                                & $256$      		   \\ \midrule
    Latent vector dimension ($[d_1,d_2]$).				  & $[20, 5]$      	   \\
    Langevin Dynamics steps during training.			  	  & $25$      	   	   \\
    Langevin Dynamics steps during inference.			  & $500$      	   	   \\
    Langevin Dynamics step size ($s_z$).					  & $0.001$      	   \\
    Langevin Dynamics sigma ($\sigma_z$).				  & $0.025$         \\ \midrule
    Initial learning rate.                                & $1e^{-3}$             \\
    Training steps.                                       & $1000$             \\ 
    Batch Size.                                           & $4$                \\
    Learning rate decay (3 times).                        & $0.8$              \\ \bottomrule
    \end{tabular}
\end{table}

\section{Time-series forecasting}

\begin{figure}[ht!]
\centering
\includegraphics[width=\linewidth]{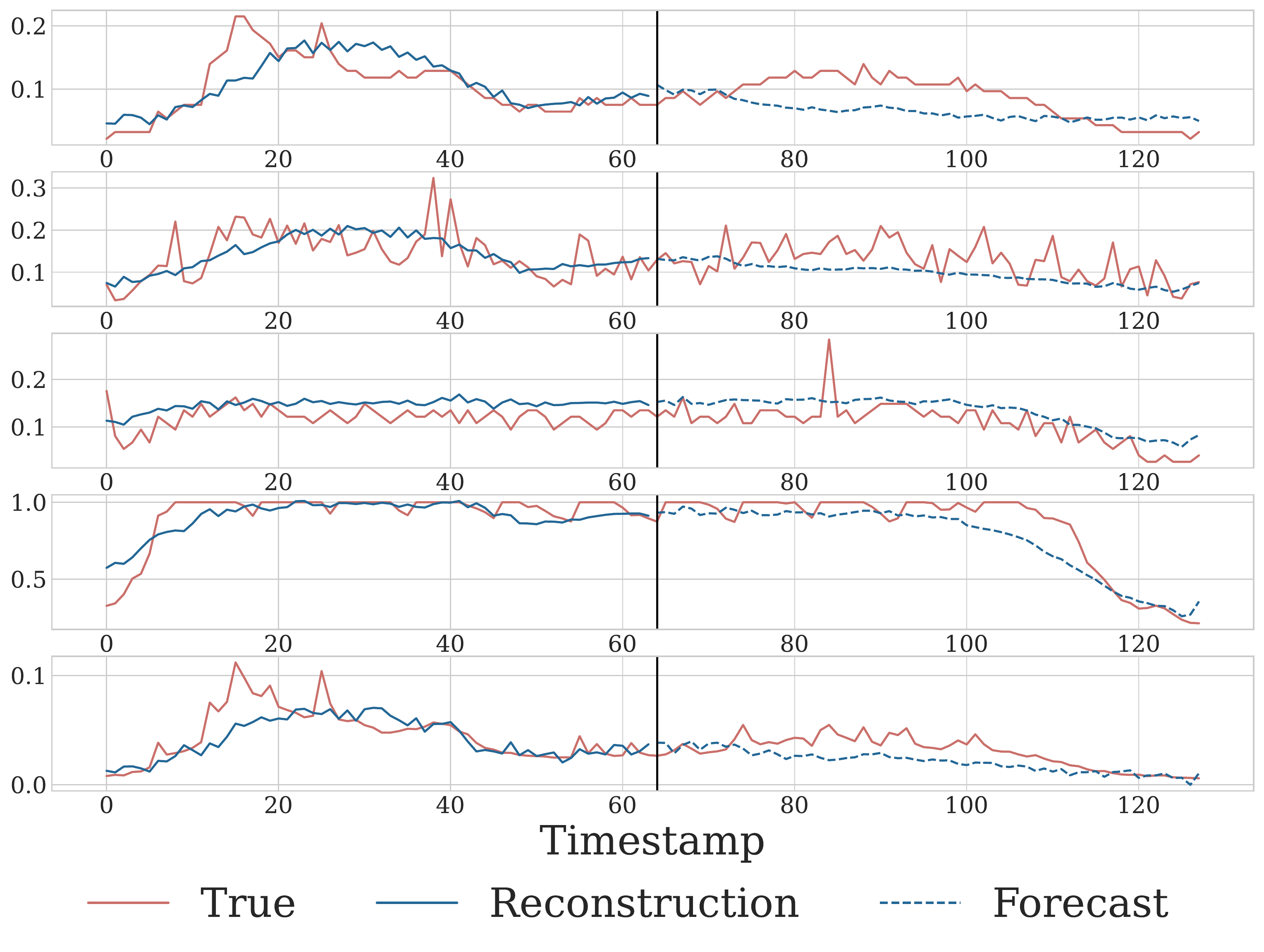} 
\caption{Example of forecasts produced by \OurModelsp on a window of machine-1-1 test set of SMD. The model is trained to generate windows of size 128, the first 64 timestamps are available during inference of latent vectors, and the last 64 correspond to the forecasts.}
\label{fig:forecasting}
\end{figure}

\OurModelsp can be used as a forecasting model without any changes to the training procedure or architecture. \OurModelsp can forecast future values by masking them during the inference of latent vectors, analogous to how the model handles missing data. The observed timestamps (left to the forecasting starting timestamp as represented with the vertical line of Figure \ref{fig:forecasting}) are used to infer the current latent vector, which is used to generate the whole window, producing the forecasts.

Figure \ref{fig:forecasting} shows an example of the forecasts produced for a subset of machine-1-1 of the SMD dataset. In this example, the model is trained to reconstruct and generate windows of size 128. The first 64 timestamps are available during the inference of the latent vector, and the last 64 corresponds to the forecasted region.

\end{document}